# Assessing the Reach and Impact of Game-Based Learning Approaches to Cultural Competency and Behavioural Change

Ian Dunwell, Panagiotis Petridis, Petros Lameras, Maurice Hendrix, and Stella Doukianou
Serious Games Institute
Coventry University
United Kingdom
+44(0)2476887688
idunwell@cad.coventry.ac.uk

Mark Gaved
Institute of Educational Technology
The Open University
United Kingdom
2nd line of address
+44(0)1908654821
mark.gaved@open.ac.uk

## ABSTRACT
As digital games continue to be explored as solutions to educational and behavioural challenges, the need for evaluation methodologies which support both the unique nature of the format and the need for comparison with other approaches continues to increase. In this workshop paper, a range of challenges are described related specifically to the case of cultural learning using digital games, in terms of how it may best be assessed, understood, and sustained through an iterative process supported by research. An evaluation framework is proposed, identifying metrics for reach and impact and their associated challenges, as well as presenting ethical considerations and the means to utilize evaluation outcomes within an iterative cycle, and to provide feedback to learners. Presenting as a case study a serious game from the Mobile Assistance for Social Inclusion and Empowerment of Immigrants with Persuasive Learning Technologies and Social Networks (MASELTOV) project, the use of the framework in the context of an integrative project is discussed, with emphasis on the need to view game-based learning as a blended component of the cultural learning process, rather than a standalone solution. The particular case of mobile gaming is also considered within this case study, providing a platform by which to deliver and update content in response to evaluation outcomes. Discussion reflects upon the general challenges related to the assessment of cultural learning, and behavioural change in more general terms, suggesting future work should address the need to provide sustainable, research-driven platforms for game-based learning content.

## Categories and Subject Descriptors
I.3.8 [Computer Graphics]: Applications; K.3 [Computers and Education]: Computer Uses in Education—Computer-assisted instruction; K.4 [Computers and Society]: Social Issues—Employment

## General Terms
Design, Human Factors

## Keywords
Game-based learning; Serious Games; Inclusivity; Cultural learning; Mobile learning

## 1. INTRODUCTION
Game-based learning has been shown to be effective in a wide range of contexts, including cultural learning [1]. However, as with any instructional medium, differences in audience, context, and representational medium each demand individual design considerations. This in turn creates a challenge in translating research outcomes, which typically analyze the efficacy of an individual game, into generalisable findings capable of feeding in to future designs. In Section 2, a range of evaluation methods applied to digital games for learning are described, alongside the challenges specific to cultural competency development. This forms the basis for the framework proposed in Section 3, which seeks to reconcile the unique nature of gameplay as a form of data capture with the need to provide validated research outcomes which allow for comparison to other forms of education. An application of this framework to the MASELTOV project is described in Section 4, alongside an overview of the developed serious game and its role within the project. An important observation is the need to consider the game both as a standalone entity with measurable learning outcomes, and as part of a wider learning process supported by resources and tools across the project.

## 2. BACKGROUND
Assessment of the overall efficacy of an intervention can prove challenging in a cultural learning context. If we seek to simply transfer knowledge, then summative assessment may provide a metric against which to evaluate the efficacy of an intervention. However, factual knowledge in itself does not necessarily equate to attitudinal or behavioural change. Attempts to incorporate competency assessment in a game-based context have noted the high resource demands in creating competency metrics, though their reusability is of value [2]. Such metrics for cultural learning frequently focus on specific cultures and contexts, rather than providing a ubiquitous solution; however, generalizable themes exist: cultural competence is not about an individual being able to emulate another culture, rather, it is about their capability to recognize cultural differences and respond effectively. Knowledge is an important first step in this process, but this knowledge must also be translated to behaviour. This can be perceived as a three-step process in education: knowledge gain results in attitudinal change, which subsequently influences behaviour. Each step poses both educational and evaluative challenges, since the causal chain includes many cofactors. Consider, for example, the case of road crossing safety: knowledge in this case is a simple rule set of actions required to cross safely, and can be directly assessed. Attitude presents a more complex challenge, as self-reporting of intended safe behaviour may not necessarily equate to safe behaviour in practice [3]. Similarly, cultural competence may be translated into a set of transferrable understandings, though this requires the individual be motivated to apply these in practice.

A systematic review of literature on the efficacy of serious games noted the diverse range of evaluation methodologies and techniques used, in part a consequence of the diverse range of topic areas and fields to which games have been applied [4]. It noted the need for objective, quantitative evidence, such as that generated by randomized control trials, to gain an ideal metric of efficacy. The challenge in practice, however, is that generating such evidence as part of an iterative development cycle is seldom viable, due to the time and costs involved. Furthermore, whilst an approach which utilizes a randomized-control trial may show efficacy when comparing game and control, it generates less insight into how the design might be refined. Consequently, smaller-scale qualitative studies are frequently applied at the design stage to gain this insight in a pragmatic fashion. A further approach to quantitative evaluation specific to the mobile context is presented through Quantitative Evaluation Framework (QEF) [5]. The QEF emphasizes factors in evaluation such as the individual who conducts it, its timing relative to the project, and the purpose of the evaluation. This purpose is a particularly important consideration; as noted an evaluation seeking to feed-in to design may benefit less from a quantitative approach than an overall evaluation of efficacy.

Mobile devices have advantages as platforms for game-based learning, being able, with consent, to report a users location alongside other data which may have relevance at the evaluation stage. Another quantitative evaluation approach considers the technology acceptance model as its basis [6], building upon the notion of uptake resulting from a combination of perceived usefulness and usability. "Usefulness" in the case of game-based learning can prove challenging to define, as a game may be perceived as both a useful form of entertainment, or useful educational resource. Interpreting users' self-reporting, therefore, requires caution in understanding the distinction between these two perceptions, and their implications for game design. Provided this can be ascertained, this can prove a useful tool for establishing the required balance between entertainment and educational aspects.

Considering qualitative aspects in more detail, understanding learner experience can also be of value, with an experience design standpoint having been applied to guide the balance between "positive" and "serious" aspects of a game [7]. Given the need to balance education and entertainment carefully during the development of a game-based learning solution [8], experience design can offer an approach to understanding both the engaging and educational aspects of a game from an end-user perspective. Bocconi et al. [9] note the significance of the "threshold concept" in competency learning, suggesting key transformative points can be defined which result in a step-change in understanding or behaviour for the learner. If such concepts can be identified, then the evaluation process can be simplified by focusing on assessing the learner's understanding of these key concepts, rather than attempting a broader assessment of knowledge. From an educational perspective, evaluation frameworks have noted the need to understand the serious gaming experience from the learner's standpoint, in order to gain a perspective on learning outcomes not immediately relatable to assessment metrics [10].

Pandeliev and Becker propose a framework for online evaluation [11], noting the difficulties in applying in-person experimental protocols. This leads to a significant challenge in the assessment of any online tool seeking to achieve attitudinal or behavioural change, including serious games: Assessment in an online context may allow us to work directly with end-users, though it is unlikely it will facilitate the in-depth understanding which could be achieved through qualitative techniques such as interview or case study. However, it does hold advantages: we are no longer constrained to working out-of-context to evaluate serious games in a laboratory, rather, we are doing research "in the wild" with an online audience. However, numerous considerations emerge when shifting from laboratory evaluation to a real end-user base. Firstly, research is contingent on participants, and this implies the game be capable of attracting and retaining an audience. In any case, this would be the first metric of success, however, assuming an iterative approach is adopted, it has a fundamental consequence for game-based learning: games must get the fun, engaging aspects correct, before researchers can iteratively evaluate instructional efficacy and progress the game's design. Without the capacity to attract and retain users, assessing the educational impact in the wild is impossible.

Given the potential of community formation as a basis for social learning, an argument exists from this perspective to focus firstly on creating an entertainment game and fostering a user-base, then working with this user base through a process such as participatory design to create an effective learning experience. A review of assessment techniques in serious games [12] notes key future research goals to be characterization of players' activity, and better integration of assessment within games. When working with an active end-user base, assessment can provide useful feedback to both the developer of the game, and the learner themselves. In both cases, data from the same source can be collected, and it is the interpretation and presentation of this information that defines it as design input or feedback to the learner.

The application of technologies such as gaze-tracking have shown themselves to be promising metrics for evaluation. A study examining gaze in the specific case of a serious game showed not only a difference between genders, but also the ability to correlate the metric to overall performance within the game [13]. Admittedly, the serious task in this case was not related to behavioural or attitudinal change, and applying this or similar biometrics has limitations in terms of the type of learning a game is seeking to convey, though it remains an interesting area for future work.

An recurrent theme in serious game development is the need for collaboration between a wide range of stakeholders through the development process. Developing digital games requires a game and pedagogical designer: rarely does a single individual possess both skill-sets, and the "tension" [14] between game and instruction requires careful negotiation between the two fields. Furthermore, stakeholders include members of the target audience, subject matter experts, and researchers, as well as the individuals responsible for translating a design concept into a functional prototype, which requires both artistic and technical skills. Participatory design is often advocated, but difficult to implement [15], in particular as individuals can often report subjectively if a game is "fun", though self-reporting of learning, and in particular impact on attitude, can deviate significantly from objective observations of resultant behaviour [16].

In summary, the challenge in evaluating the efficacy of a digital game seeking to teach cultural competencies is analogous to any other game-based approach seeking attitudinal and behavioural change: simply transferring knowledge does not guarantee this change, and therefore a risk exists in using knowledge alone as a proxy for impact. Cultural competency development also poses some unique questions for researchers and developers of game-based solutions: as we cannot hope to represent the breadth and depth of real-world cultures within the confines of a digital game, how can we distil key principles and identify the "thresholds" [9] at which competencies develop? Can we rely on self-reporting of competences and participatory techniques when seeking to engage in an iterative design process, given the differences in perceptions of usefulness - as game or educational tool - which might emerge from the target audience? This section has presented a range of quantitative and qualitative frameworks. In the following section, we relate these to the specific case of cultural competence development, proposing an evaluation framework applied within the case study presented in Section 4.

## 3. ASSESSMENT FRAMEWORK

Reflecting on the background of the previous section, this Section proposes an assessment approach for digital games applied for the development of cultural competences. Firstly, the two key objectives common to interventions of reach and impact are described. Reach is considered in terms of game deployment and uptake, whilst Impact seeks to consolidate the diverse range of techniques identified in the previous section to form a basis on which to propose clear avenues suited to both the development process of serious games, and the need to demonstrate efficacy. In doing so, we consider two key evaluation contexts: laboratory study, and deployment with end-users. The latter raises a number of important ethical considerations, principally surrounding informed consent, noted in Section 3.3. The subsequent question, given that metrics of reach and impact can be established, is how to best feed the outcomes of research back to both game designers, to support iterative development, and learners themselves, in the form of feedback and progress indicators, discussed through case study and conclusions in Section 5.

### 3.1 Reach

When deployed in an online marketplace, uptake can be measured in terms of number of users, with techniques such as IP geolocation or location awareness applied to gain high-level insight into the demographic. For games targeting general awareness raising this may be sufficient; however, in many cases games seek to target excluded or disengaged demographics. Evaluation in a laboratory context is unlikely to be representative of these groups, as they are less likely to engage with the research process or opt-in to studies. Survey of the player-base is a potential route to gain further understanding of reach, and can be incentivized through in-game rewards such as content or collectables. Correlation, for example, of reported postcode to indices of multiple deprivation can offer a means to gain an understanding of the socioeconomic background of players without requiring explicitly surveying, though must be conducted in an ethical fashion following the principles outlined in Section 3.3. In the case of a game seeking to enable immigrants to develop their cultural competences, such as that described in Section 4, it may also be possible to correlate overall population statistics to total numbers of users, allowing for approximations of reach to target audiences.

### 3.2 Impact

Section 2 illustrates the diverse range of methods by which the impact of digital games for learning has been assessed. Common to many of these methods is a clear distinction between evaluations performed at the development stage, which seek to gain understanding into how the impact of the game might be improved or refined, and those undertaken post-hoc to assess the overall impact in-depth. In the case of the former, qualitative studies using methodologies such as focus groups or case studies have been shown to provide a sound basis for eliciting user-feedback and providing feed-in to the design process. Participatory design is equally valuable, though an objective view of participants' contributions is essential, as individuals willing to form part of a participatory game-design process many not be representative of the target end-user base.

Two differing design-phase approaches to sourcing participants exist: either laboratory-based evaluations, which recruit participants through standard experimental protocols, or research with the emerging player community for the game. The latter has potential benefits; involvement in the development of the game itself can stimulate engagement, and result in learning amongst participants through the process of making, as well as playing the developing game. Returning to the need for balance between entertainment and education noted in Section 2, this does, however, place pre-requisites on the game's design and development: it must be developed in a form suitable for engaging and retaining a community in the participatory development process, and therefore requires the entertainment aspect be brought to the forefront in early stages of development. During the development stage, assessment of impact may be broken down into a subset of requirements for the game. This is advantageous for a game's designer, as feedback from user testing can be transposed more easily to adaptations of individual game elements, rather than reporting on the game as a whole.

Following the iterative development process, a clear point should be defined at which to progress to a more detailed assessment of efficacy using a methodology such as a randomized control trial. This need not serve as an end-point to the development process, rather, it should be made explicit - in view of the impact of purpose on study design noted in Section 2 - that the goal of the evaluation is to provide a clear assessment of the overall efficacy of the game-based approach. Again, the comparison here is between performing this evaluation using an active end-user base of the game, or working within a laboratory context. Previous meta-analyses have advocated the use of randomized control trials as post-hoc evaluations [4], though challenges exist in their implementation. What, for example, can serve as a control for a serious game? Comparing versus no intervention can have limited value, unless clear quantitative metrics can demonstrate the extent of impact, or the methodology allows for clear and direct comparison to other interventions or training programmes. One option is to compare differing versions of the game, which may generate useful information for subsequent development. Though games make be a unique case in some respects, as part of an overall learning programme, methods used to validate and evaluate previous learning approaches may also hold merit.

Developed metrics for cultural competencies can be applied to create a comparable assessment of impact.

The final consideration, therefore, is how these metrics might be applied in both the real-world and laboratory context. Self-reporting of competency, for example, may vary between these two contexts, as participants directly communicating with the research develop bias towards positive reporting when compared with online users detached from the research process. Where practical, therefore, a combined approach seeking to gain in-depth insight in a laboratory context, coupled to larger-scale data collection from the game's playerbase, has obvious merit.

### 3.3 Ethical Considerations

Whilst ethical principles and practices can be adhered to in a laboratory setting with relative ease, as the researcher has direct contact with participants to ensure informed consent, research with online audiences can prove a more complex issue. Consent information can be overlooked or "clicked-through" in the desire for the player to experience the game, thus, a design requirement is that consent should be given in an opt-in fashion, with the player able to fully review the implications of their consent. As games may be intrinsically incentivizing to play, a risk also exists of players consenting to gain access to the game, rather than in an informed and ethical manner. Hence, a further advocation is that access to the game should be provided irrespective of a player's consent to participate in the research process. In the case of cultural learning and competency development, members of the target audience may also possess limited language skills, and consent approaches should demonstrate awareness of, and support for, communication of research objectives and the implications of participation to these players.

### 3.4 Implementation

Some final consideration should be afforded to how the nature of a game's design and deployment can contribute to the application of the metrics identified in Sections 3.1 and 3.2. Content distribution platforms, such as Google Play, are increasingly allowing developers to not only reach wide audiences, but also gain insight into their behaviour through the integration of metrics within the game engine. Noting the ethical considerations in Section 3.3, care should be taken to ensure informed consent is sought, but with such mechanisms in place it becomes possible to quickly ascertain metrics of reach, such as the geographical distribution of players, and impact, through measures such as playtime. However, a more concrete demonstration of impact in educational or behavioural terms can also require participants self-report or agree for their performance to be monitored. Here in particular mobile platforms can serve a useful purpose in allowing participants to self-report data on the move, and games can provide a means for incentivizing and sustaining involvement in the research process, for example rewarding participation in surveys with in-game items.

## 4. CASE STUDY I: MASELTOV

The Mobile Assistance for Social Inclusion and Empowerment of Immigrants with Persuasive Learning Technologies and Social Networks (MASELTOV) project seeks to apply a range of services to support the integration of immigrants entering Europe. Game-based learning is deployed within the project with the aim of reaching audiences who may not respond to more formal learning resources, and to convey cultural learning in an engaging and immersive form. The use of mobile technologies within the MASELTOV project is reflected in the game's use of the Android platform as a basis for deployment.

To address the challenge of providing cultural learning both within the context of the project, and in a form suitable for deployment on Google Play, several themes have been considered. The game seeks to empower the user through a fictional narrative which places them as a hero seeking to reconcile a "dimensional split", in which reality has divided into two discrete dimensions, each with a distinct culture. Hence, they are required to traverse the two dimensions as they move through the game world, experiencing cultural differences first-hand. Dialogic interactions are used extensively, with users having the freedom to choose a variety of responses to situations and experience the consequences. Success requires they develop their understanding of each of the two cultures, and apply this understanding to problem-solve. The narrative of the game takes the player through themes including travel, jobseeking, healthcare, and shopping, with problems faced derived from discussion with non-governmental organizations (NGOs) working with immigrants on a daily basis.

Empowerment is a common and demonstrably effective approach used in games seeking to influence behaviour [17]. It appears particularly suited to the case of a game for immigrants, as cultural exclusion can be seen to be linked to disempowerment: excluded immigrants feel they have no role in influencing their host country's attitudes, policies, and systems. Following the theme of empowerment, we adopt an approach taken by other serious games, which combines a partially-abstracted narrative together with an overarching story seeking to both reflect common challenges faced by immigrants, whilst presenting this from a position of empowerment.

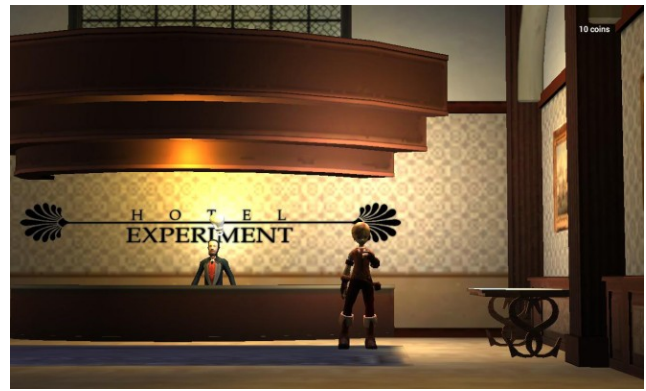

**Figure 1: Screenshot from the MASELTOV game. Avatars and dialogues are used to convey cultural learning content.**

This potential has been reflected in a number of studies seeking to utilise and understand the role a game might play in empowering an individual or community. An example of their use to empower hospital patients showed efficacy in a real-world context [17], supporting the view that games can empower the player through a range of mechanisms. Narrative is one means by which to achieve this [18], as characterization and identification can be utilised as tools by which to transpose real-world problems arising from lack of empowerment, to ones which have identical traits, but are viewed from a different perspective. Consider, for example, the case of an immigrant seeking work - in a real-world context, they may feel disempowered, with little control over whether they achieve success. In a game, however, success may be granted when pedagogical objectives have been achieved, and the simplification of the process itself to focus on key learning objectives lends itself to a format in which the individual has greater control over the outcome.

Abstraction is used within the game to allow a degree of separation from real-world simulation. The world itself is fictional, as are the two distinct cultures, however, the themes and problems faced by the player are grounded in reality. This circumvents the difficulty in accurately simulating intercultural interactions, as well as criticisms arising from attempts to describe the behaviour of an individual real-world culture in specific or superficial terms.

Finally, the game seeks to apply an experiential learning paradigm, a common approach in game-based solutions [19, 20]. Coupled with abstraction and empowerment, the game seeks to allow the user to experience situations and learn through their actions and responses. An in-game journal system scaffolds the reflection process by updating the player with their characters observations and conclusions regarding cultural differences and how best to approach situations. This in turn seeks to link to additional resources within the project which provide more formal or specific learning materials. Kolb's theory of learning through experience [21] has seen much attention from game- and simulation-based learning communities [19, 20, 22]. It has clear parallels related to how games can allow players to explore problems, devise solutions, and observe the consequences of their actions.

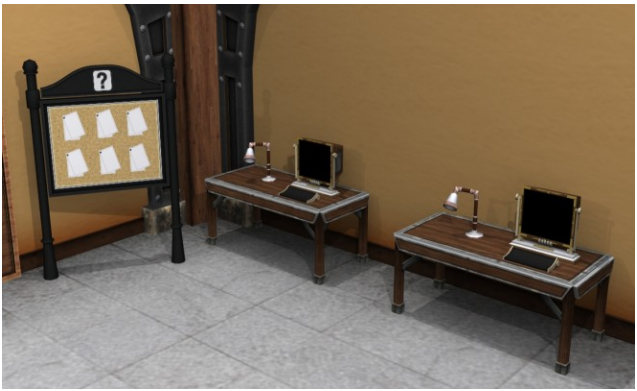

**Figure 2: Areas such as the job centre allow players to observe the differences between cultures in the job application process.**

By allowing the player to experience situations in which cultural differences present themselves as difficulties, and guiding them towards solutions, a platform for experiential learning might be created. Some cautions arising from Kolb's theory are required, however: firstly, the notion of the "intuitive" learner put forward by Kolb and intrinsically related to the experiential cycle has consequences in terms of how we might assess and feed-back to learners. The intuitive learner, by this definition, takes an exploratory approach to learning, exploring worst as well as best cases: for example, when confronted by a dialogue choice, they may deliberately answer incorrectly or inappropriately to explore the outcome. This in turn means attempting to assess competence by the "correctness" of actions in game has limited value, and consequently seeking to feed-back to a learner their errors may be met with resistance or negation.

Hence, the effective application of the experiential cycle in-game requires the development of branching or open-ended scenarios, that allow the intuitive learner to play-through and explore multiple outcomes and possibilities. Another trait of this form of learner is their tendency to return to scenarios multiple times to observe different outcomes, and games can support this through additional play-thoughs. End-user testing on-site at NGOs has been utilized to gain early qualitative insight into the design of the game and immigrants' responses. Usability has been a key focus of this study, particularly as the immersive environment delivered by the game must be coupled with an interaction paradigm which is sufficiently intuitive for a wide range of levels of gaming experience. Feedback identified themes such as navigation to be central considerations in development, with the need to be able to intuitively and rapidly negotiate the game world having implications for the interface design.

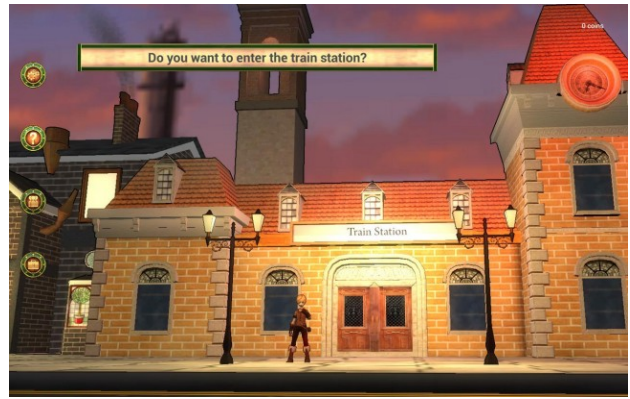

**Figure 3: Environments in the game reflect the real world, though in an abstracted form.**

A limitation of this on-site testing, however, is that the users are self-selecting and already engaged with NGOs. As the game seeks to reach a wider audience, in particular immigrants who may not be aware of or utilize NGO or online cultural learning resources, gaining representative insight into this segment of the target demographic is a key goal of future research. Deployment of the game via Google Play will be coupled with the considerations presented in Section 3, towards providing a means for understanding the audience responding to the game, and its subsequent impact. This impact may either be in direct terms of cultural learning outcomes, self-reported by players through survey incentivized via in-game reward, or objective metrics such as the proportion of players who transition to using other MASELTOV services and learning content.

## 5. CASE STUDY II: THE BEHAVIOURAL ECONOMICS OF ENERGY USE

The challenge of shaping individuals' behaviours is vital in tackling many social problems, including environmental concerns. Citizens may, for example, find it difficult to think that their individual behaviour has a meaningful impact on large-scale problems such as loss of energy sources and carbon emissions. Even though people may express environmental awareness, this is not always translated to environmentally-friendly way of life. Human behaviour is multifaceted, and driven by innumerable factors such as genetic predisposition, socioeconomic status and education. These factors interact, giving rise to vast amount of different traits. Nevertheless, social scientists have proposed various behavioural Theories and models that can take into account these factors and predict how they can influence behaviour. Hildebrand Technology Ltd, a consultancy company is currently working on a project which will help citizens compare and reduce their energy, incorporating these behavioural theories in serious games. The purpose is to shape consumers' attitude towards environmentally-conscious behaviour, through comparison, education and feedback. The project *Game Mechanics as Motivators: The Behavioural Economics of Energy Use* proposes to link a serious game with energy sensors within the participants' houses. Games driven by real-time data seek to offer a simulating environment where users will complete challenges and goals towards a positive impact on their behaviour. The game will allow users not only to stay informed about their own energy consumption, but also to see if they are high, medium or low energy users, compared to others in the same category.

A serious game which targets the general public about a change of energy usage needs to be based on predictive models. These predictive models enable the designer to create different scenarios inside the game, adapted to human's attributes and hence increasing their influence on behaviour. Serious games can become an easy and effective method of collecting the data so as to create these models. Thus far within the project, two games are being created, prior to the game for energy, in order to examine two aspects of human behaviour: risk tolerance and patience. Risk tolerance has concerned psychologists, economists, social scientists and marketers a long time ago, since people's attitude towards risk can be used potentially to encourage them to eat healthier, exercise more and even consume less energy [23]. The game related to risk-taking is based on a Lottery Behavioural Task [24] and will classify people according to their perception towards the changes in probability. This classification will allow us to measure the degree of risk aversion. Regarding the second important personality trait, patience, is more complex than risk tolerance since it consists of multiple dimensions [25]. There are different kinds of patience. In this project, patience will be referred as the perception of the immediate pay offs to larger more distant ones. The tendency of people to choose smaller immediate rewards to larger but more delayed is connected not only with the proneness to addiction [26] but also with environmental action.

A foremost challenge in this project is understanding the reach and impact of game-based intervention on end-users' behaviours. Hence, the project seeks to foster engagement with participants that will follow up and commit to the experiments, in an "in the wild" context, as described in Section 3.2. The games for patience and risk measure two traits that are notoriously difficult to capture in small-scale experiments and no matter the quality of them there is always the possibility to misinterpret the results. Moreover, risk and patience are general traits that sometimes are coupled with certain activities and it can become complex to design an ideal behavioural task that will not include bias. Consequently, a sufficient sample of real-world end users is required to gain insight into the impact of interventions.

## 6. CONCLUSIONS

Games provide a rich source of data on user interaction. Their use as research instruments is well-documented [27], though the need described in Section 2 to better integrate the assessment process into the game [12] represents only one potential approach. An alternate technique is to record information in a form suitable for externalized analysis. This requires rich data capture be implemented into the game, recording in detail user interactions ranging from individual touch interactions on a mobile device, though to higher-level information such as content preferences and engagement times. This has appeal from a research perspective, as it has the potential to open up access to data captured during gameplay to a wide range of analysis tools and techniques. In such an approach, the requirement is not that the assessment is incorporated into the game as a whole, rather, that the externalized data can be used to generate feedback communicated either through the game itself, or a wider, blended learning approach.

In the case of integrative approaches combining multiple tools, such as that of MASELTOV presented in Section 4, a blended approach to learning which seeks to support learners in traversing a wide range of tools and resources requires that the impact of the game be consider and evaluable in wider terms. Games may provide a valuable starting point for learners, particularly where desired learning outcomes require they first engage and acquire intrinsic motivation to learn, rather than requiring extrinsic motivators. Whilst evaluating and understanding individual components of a blended learning experience can guide development, assessing the joint impact of these components and their interplay is also frequently required.

In conclusion, a clear need exists for two distinctions in the evaluation process for game based learning approaches to cultural competence development. The first is the distinction between evaluations which seek to gain insight into the behaviour of the players of the deployed game (or prototype), compared to evaluations which take place in a laboratory context. Both have merits; the former consists of a representative sample, and can provide valuable information into the impact of the game on its target audience, the latter allows researchers greater access to participants and hence more qualitative insight.

## 7. ACKNOWLEDGMENTS

This work has been supported by the European Commission under the Collaborative Project MASELTOV ("Mobile Assistance for Social Inclusion and Empowerment of Immigrants with Persuasive Learning Technologies and Social Network Services") funded by the European Commission under the eInclusion theme, project FP7-ICT-7 Grant agreement n. 288587.